%% file: acl_latex.tex
\documentclass[11pt]{article}
\usepackage{makecell}
\usepackage[preprint]{acl}
\usepackage{amsmath}
\usepackage{booktabs}
\usepackage{multirow}
\usepackage{amsfonts}
\usepackage{times}
\usepackage[dvipsnames]{xcolor}
\usepackage{latexsym}
\usepackage{graphicx}
\usepackage{subcaption}
\usepackage{xcolor} 
\usepackage[most]{tcolorbox}
\usepackage[T1]{fontenc}
\usepackage[utf8]{inputenc}

\usepackage{microtype}

\usepackage{inconsolata}

\usepackage{graphicx}

\title{Reasoning through Exploration: A Reinforcement Learning Framework for Robust Function Calling}

\author{
 \textbf{Bingguang Hao\textsuperscript{1}$^{\ast\dagger}$},
 \textbf{ZengZhuang Xu\textsuperscript{1}$^{\ast}$},
 \textbf{Maolin Wang\textsuperscript{2}$^{\ast\dagger}$},
 \textbf{Yuntao Wen\textsuperscript{1}$^{\ast\dagger}$},
 \textbf{Yicheng Chen\textsuperscript{1}$^{\dagger}$},
 \\
 \textbf{Cunyin Peng\textsuperscript{1}},
 \textbf{Long Chen\textsuperscript{1}},
 \textbf{Dong Wang\textsuperscript{1}},
 \textbf{Xiangyu Zhao\textsuperscript{2}},
 \textbf{Jinjie Gu\textsuperscript{1}},
 \textbf{Chenyi Zhuang\textsuperscript{1}},
 \textbf{Ji Zhang\textsuperscript{3}}
\\
 \textsuperscript{1}\includegraphics[height=1em]{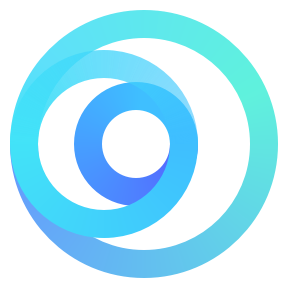}\,AWorld Team, Inclusion AI \quad
  \textsuperscript{2}City University of Hong Kong \quad
 \textsuperscript{3}Southwest Jiaotong University \\
 \texttt{\{bingguanghao7,jizhang.jim\}@gmail.com}\quad
 \texttt{\{chenyi.zcy\}@antgroup.com}
}

\begin{document}
\maketitle
{\renewcommand\thefootnote{}
\footnotetext{$^\ast$Equal contributions. $^\dagger$Work done at Ant Group.}
\footnotetext{\ \ Project Page: \href{https://github.com/BingguangHao/RLFC}{EGPO}}
\begin{abstract}


The effective training of Large Language Models (LLMs) for function calling faces a critical challenge: balancing exploration of complex reasoning paths with stable policy optimization. Standard methods like Supervised Fine-Tuning (SFT) fail to instill robust reasoning, and traditional Reinforcement Learning (RL) struggles with inefficient exploration. We propose \textbf{EGPO}, a new RL framework built upon Group Relative Policy Optimization (GRPO), designed to address this challenge directly. The core of EGPO is an entropy-enhanced advantage function that integrates the entropy of the model's Chain-of-Thought (CoT) into the policy gradient computation. This encourages the generation of diverse reasoning strategies. To maintain optimization direction, the entropy bonus is carefully constrained by a clipping mechanism. Complemented by a strict, binary reward signal, EGPO effectively guides the model towards discovering structured and accurate tool invocation patterns. 
On the challenging Berkeley Function Calling Leaderboard (BFCL), a 4B-parameter model trained with EGPO sets a new state-of-the-art among models of comparable size, surpassing a range of strong competitors, including GPT-4o and Gemini-2.5.
\end{abstract}

\input{sections/intro}

\input{sections/related}
\input{sections/pre}
\input{sections/method}
\input{sections/experiment}


\section{Conclusion}


This work presents EGPO, a novel and effective RL framework for LLM function calling. By encouraging diverse reasoning paths while maintaining optimization stability, EGPO enables models to develop more sophisticated parameter extraction and verification strategies. EGPO achieves state-of-the-art results among similar sized models and outperforms a series of strong competitors like GPT-4o and Gemini-2.5 with only 4B parameters, representing a significant advancement towards building more intelligent and reliable LLMs for real-world tool interaction.

\section*{Limitations}
While EGPO demonstrates strong performance on multiple function calling benchmarks, two key limitations remain. First, EGPO relies on high-quality CoT reasoning during training, which may not generalize effectively to domains where structured reasoning is less natural or poorly defined. Second, the stringent binary reward design, though effective in promoting format correctness and semantic accuracy, may be too rigid for tasks requiring nuanced partial credit or multi-step reasoning with intermediate feedback. 

\bibliography{custom}

\appendix

\input{sections/appendix}

\end{document}

%% file: sections/intro.tex
\section{Introduction}

Function calling represents a pivotal advancement in the evolution of Large Language Models (LLMs), transforming them from text generators into practical, interactive agents capable of addressing real-world challenges~\cite{wang2025function,qu2025tool}. This capability is critical as it bridges the gap between an LLM's vast internal knowledge and external, dynamic resources, thereby significantly enhancing its functionality, accuracy, and overall utility~\cite{patil2024gorilla,shen2024llm}. By enabling LLMs to interact with APIs, execute code, or access up-to-date information, function calling allows them to transcend the limitations of their static training data~\cite{nguyen2024sfr}.


\begin{figure}[t!]
\centering
\includegraphics[width=1.0\columnwidth]{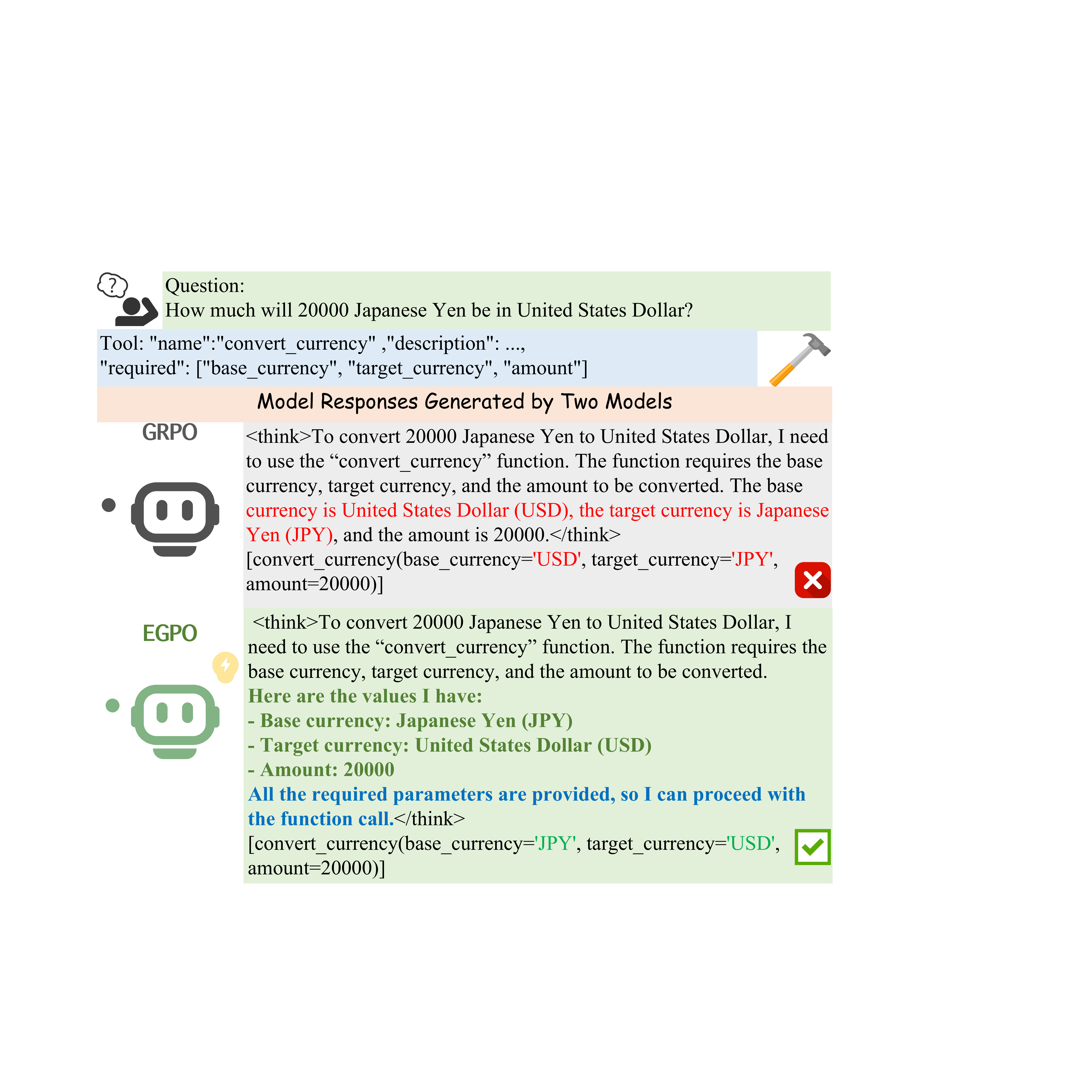} 
\caption{By employing a more structured thinking and verification process, our EGPO framework correctly identifies all required parameters for the tool call, whereas the baseline GRPO model fails by swapping the arguments.}
\label{fig:think}
\end{figure}

However, the effective training of LLMs for function calling faces a central challenge: balancing the exploration of complex reasoning paths with stable policy optimization. Prevailing training methodologies, namely Supervised Fine-Tuning (SFT)~\cite{hao2025funreason,liu2024toolace} and Reinforcement Learning (RL)~\cite{qian2025toolrl,zhang2025nemotron}, struggle to resolve this tension. SFT, which relies on mimicking distilled expert trajectories, often leads to superficial pattern matching rather than instilling robust reasoning capabilities~\cite{setlur2025e3,qian2025toolrl}. While effective for format adherence, it fails to equip models with the adaptability required for diverse, unseen scenarios~\cite{zeng2025toolace,prabhakar2025apigen}. 

Reinforcement learning (RL), though theoretically suited for exploration, encounters significant obstacles in this domain~\cite{qian2025toolrl}. First, the problem of Signal Scarcity is acute; the structured and syntactically strict nature of function calls means that random exploration rarely yields a valid Chain-of-Thought (CoT), depriving the model of a meaningful learning signal~\cite{gao2024designing}. Second, for base models that already possess some degree of tool-using ability, Rewarding Redundancy becomes a critical issue~\cite{zhang2025nemotron}. Overly complex, process-based rewards often lead to suboptimal results, failing to refine the model's performance effectively~\cite{guo2025deepseek}. The key to advancing the field is not just generating correct function calls, but effectively guiding the model to develop sound and generalizable thinking patterns~\cite{team2025kimik2}.

In this work, we introduce \textbf{EGPO}\textbf{(Entropy-Guided Policy Optimization)}, a novel RL framework designed to foster strategic and diverse reasoning. Built upon Group Relative Policy Optimization (GRPO)~\cite{shao2024deepseekmath}, EGPO directly integrates the entropy of the model's Chain-of-Thought into the advantage calculation, creating a policy gradient that encourages the exploration of diverse reasoning pathways. To ensure this exploration does not derail the learning process, the entropy bonus is constrained by a carefully designed clipping mechanism, which prevents it from inverting the sign of the original advantage and thus preserves a stable optimization direction. 
Additionally, the exploration strategy is complemented by a stringent 
Single-Criteria Reward mechanism, which provides an informative binary signal: a reward is granted only when the model's output is perfect in both format and accuracy. 
By combining entropy-guided exploration with a clear, unambiguous reward, EGPO effectively guides models to discover superior and more structured thinking patterns for tool use (Figure~\ref{fig:think}). 
Our experimental results on the Berkeley Function Calling Leaderboard (BFCL)~\cite{patil2024gorilla} and other two benchmarks~\cite{li2023api,chen2025acebench} demonstrate that EGPO achieves superior performance among open-source models.

Our contributions are listed as follows:

\begin{itemize}
\item We propose \textbf{EGPO} that integrates CoT reasoning to guide policy exploration while maintaining optimization stability.
\item We show that a stringent Single-Criteria Reward is a highly effective and efficient method for improving the tool-use capabilities of proficient foundation models.
\item Our 4B model achieves state-of-the-art results among similar sized models and outperforms a series of strong competitors like GPT-4o and Gemini-2.5 on BFCLv3.
\end{itemize}

%% file: sections/related.tex
\section{Related Work}
\subsection{Function Call}
Function calling represents a pivotal advancement in the field of LLM, transcending their traditional role of mere text generation to empower them with dynamic interaction capabilities with external environments~\cite{zhang2024xlam,hao2025funreason,2025InSpire}. This paradigm shift enables LLMs to interface seamlessly with a vast array of tools, Application Programming Interfaces (APIs), and databases, thereby unlocking a new realm of possibilities~\cite{li2023api,chen2025acebench}. Through function calling, LLMs gain the ability to access real-time information, perform specific actions in the real or simulated environments, ensure the factual accuracy of their responses by consulting authoritative sources, and handle complex computations that are beyond their inherent symbolic manipulation capabilities~\cite{qin2025meta}.

\begin{figure*}[]
\centering
\includegraphics[width=\textwidth]{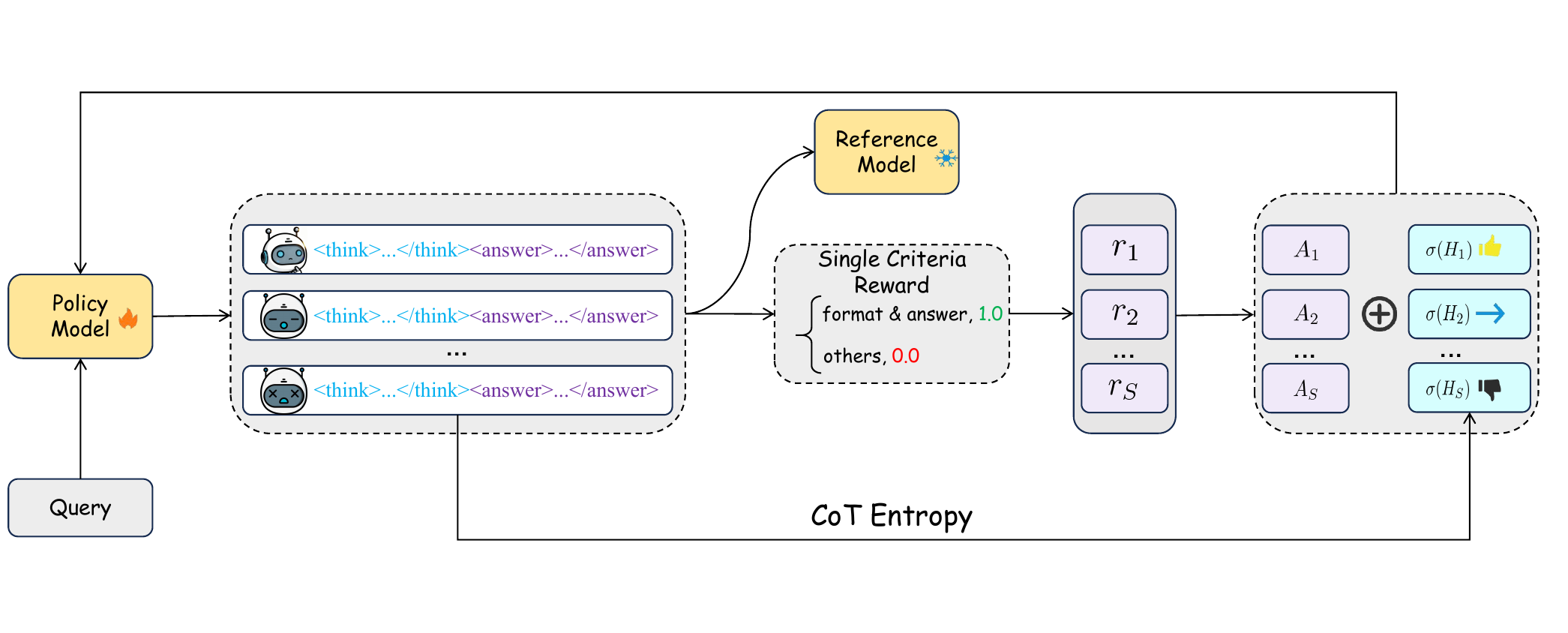} 
\caption{Overview of our EGPO framework. For a given query, EGPO calculates rewards using a single-criteria function and integrates CoT entropy with the advantage signal to guide the policy's exploration of reasoning paths.}
\label{fig:pipeline}
\end{figure*}

Various approaches have been developed to facilitate and enhance function calling~\cite{qian2025toolrl,hao2025funreason}. These range from strategic prompt engineering, where specific instructions guide the LLM to recognize and utilize functions, to fine-tuning existing LLM architectures on datasets rich with function call examples~\cite{zhang2024xlam}. More specialized architectures are also emerging that are designed from the ground up with function calling in mind. Retrieval Augmented Generation (RAG) approaches, when combined with detailed tool descriptions, enable LLMs to dynamically retrieve and employ the most appropriate tools based on the user's query~\cite{nguyen2024sfr}. Furthermore, advanced agentic frameworks are being developed that allow LLMs to engage in multi-step planning and execution, making autonomous decisions about when and how to use various tools~\cite{team2025kimik2}. However, most existing approaches primarily rely on supervised learning paradigms, which may limit the model's ability to explore diverse reasoning strategies and adapt to novel function calling scenarios~\cite{liu2024toolace}.

\subsection{LLM Reasoning with RL}

Reinforcement Learning plays a transformative role in significantly enhancing the reasoning capabilities of LLMs, moving them beyond mere statistical pattern matching to embody more robust and sophisticated cognitive functions, such as logical deduction, complex problem-solving, and strategic decision-making~\cite{guo2025deepseek,team2025kimi}. While LLMs inherently exhibit emergent reasoning abilities, RL provides a powerful framework for refining and amplifying these nascent capabilities. Algorithms like Proximal Policy Optimization~\cite{schulman2017proximal} and Group Relative Policy Optimization~\cite{shao2024deepseekmath} are commonly used to optimize the LLM's policy (its decision-making process for generating text or selecting actions) based on the received rewards~\cite{zhang2025policy}. This enables LLMs to learn from environmental feedback in a more nuanced and effective way.

For instance, an LLM can learn to strategically utilize external tools by being rewarded for successfully leveraging them to solve problems or retrieve accurate information~\cite{qian2025toolrl,zhang2025nemotron}. Similarly, RL can refine dialogue interactions, allowing LLMs to engage in more coherent, contextually aware, and goal-oriented conversations~\cite{hu2023enabling}. Beyond just improving response quality, RL helps LLMs develop a deeper understanding of task objectives and the underlying logical structure of problems~\cite{pternea2024rl,wang2024reinforcement}. Nevertheless, current RL approaches for function calling often struggle with balancing exploration and exploitation, particularly when the model needs to generate both reasoning steps and precise function calls~\cite{zhang2025nemotron}.  EGPO addresses this challenge by incorporating CoT entropy into the advantage calculation, encouraging the model to explore diverse reasoning paths while maintaining stable optimization for accurate function calling.


%% file: sections/pre.tex
\section{Preliminaries}

This section formally defines the function calling task and introduces the Group Relative Policy Optimization (GRPO) algorithm.

\subsection{Problem Definition}

Let \(q\) be a user query sampled from a dataset \(\mathcal{D}\). For each query, a set of available tools \(T = \{ t_1, \dots, t_N \}\) and the reference answer \(g\in \mathcal{G}\) are provided. In this context, the LLM is treated as a policy \(\pi\) within the reinforcement learning framework, mapping environmental states to actions. Given the state comprising the query \(q\) and tool set \(T\), the policy \(\pi\) generates a set of rollouts \(\mathcal{O} = \{o_1, \dots, o_S \}\). Each rollout \(o_i\) consists of a Chain-of-Thought (CoT) reasoning sequence followed by the function call, denoted as \(o_i = \{ c_{i_{1}}, \dots, c_{i_{W}}, f_{i_{1}}, \dots, f_{i_{L}} \}\), where \(c_{i_{1}}, \dots, c_{i_{W}}\) are tokens in the reasoning process (enclosed within \(\texttt{<think>}\) to \(\texttt{</think>}\)) and \(f_{i_{1}}, \dots, f_{i_{L}}\) are the function call tokens. Each query and its rollouts are represented by a tuple \((q, T, \mathcal{O})\), with \(q\) and \(T\) defining the state and \(\mathcal{O}\) encapsulating the actions produced by \(\pi\).

\subsection{GRPO}

Group Relative Policy Optimization (GRPO) serves as an efficient alternative to Proximal Policy Optimization (PPO), leveraging Generalized Advantage Estimation (GAE) without the need to learn a separate value function~\cite{schulman2015high,schulman2017proximal,shao2024deepseekmath}. Instead, it estimates advantages by using the average reward across multiple sampled outputs for the same query as a baseline, followed by normalization. For a set of rewards \(\{r_1, \dots, r_S\}\) corresponding to \(S\) rollouts for a given query, the advantage \(A_i\) for rollout \(o_i\) with reward \(r_i\) is computed as:

\begin{equation}
A_i = \frac{r_i - \mathrm{mean}(\{r_1, \dots, r_S\})}{\mathrm{std}(\{r_1, \dots, r_S\})}.
\label{eq:mdp_opt_grad_grpo}
\end{equation}
The policy is optimized by maximizing a clipped surrogate objective:
\begin{equation}
\begin{aligned}
&\mathcal{J} = \mathbb{E}_{q \sim \mathcal{D}, o \sim \pi_{\text{old}}(O \mid q)}\sum_{t=1}^{S} \bigg[\\
&\min\!\Big( \rho_t \hat{A}_t, \mathrm{clip}\!\Big( \rho_t, 1\!-\!\varepsilon, 1\!+\!\varepsilon \Big) \hat{A}_t \Big)\\
&- \beta \mathrm{KL}(\pi \| \pi_{\text{ref}}) \bigg],
\end{aligned}
\label{eq:grpo_loss}
\end{equation}
where \(\rho_t = \frac{\pi(o_t \mid q, o_{<t})}{\pi_{\text{old}}(o_t \mid q, o_{<t})}\) is the likelihood ratio between the current policy \(\pi\) and the old policy \(\pi_{\text{old}}\), \(\hat{A}_t\) denotes the estimated advantage, and \(\varepsilon\), \(\beta\) are hyperparameters controlling the clipping range and KL divergence penalty, respectively. This formulation promotes stable policy updates while encouraging alignment with a reference policy \(\pi_{\text{ref}}\).

%% file: sections/method.tex
\section{Method}

In this section, we describe our proposed EGPO method in detail, including the design of the reward and CoT entropy guided policy exploration. An overview of the EGPO framework is presented in Figure~\ref{fig:pipeline}.

\subsection{Reward Design}

We design a straightforward yet effective reward function to conduct reinforcement learning. For a given query $q$ with reference answer $g$, the model's generated answer $o_i$ is evaluated as follows: If $o_i$ involves a tool call, $o_i$ is deemed correct only if it is correct by performing Abstract Syntax Tree (AST) evaluation, exactly matches $g$, and adheres to the required reasoning template. Conversely, if $g$ does not involve a tool call, $o_i$ is correct only if it fails AST parsing and complies with the reasoning template. In all cases, any output that fails to follow the thinking template will be directly deemed an incorrect answer.

The reward is thus defined as:
\begin{equation}
r(o_i) = 
\begin{cases} 
1, & \text{format \& answer are correct}  \\
0, & \text{otherwise.}
\end{cases}
\end{equation}

This binary reward emphasizes the holistic integrity of the output, ensuring not only semantic accuracy but also precise structural compliance, which is critical for downstream processing. For non-tool-calling scenarios, the AST failure condition implicitly confirms that the response is appropriately textual and avoids spurious tool invocations.

\subsection{Entropy-Guided Policy Optimization (EGPO)}

To enhance exploration in the Chain-of-Thought (CoT) reasoning process within the RL framework, our EGPO method incorporates CoT entropy into the advantage calculation. For a query $q$ and a set of rollouts $\mathcal{O} = \{o_1, \dots, o_S\}$, where $o_i = \{c_{i_{1}}, \dots, c_{i_{W}}, f_{i_{1}}, \dots, f_{i_{L}}\}$ (with $c$ as CoT tokens and $f$ as final answer tokens), the vocabulary is denoted as $\mathcal{V}$ and the average CoT entropy of rollout $o_i$ is computed as:

\begin{equation}
E_i =- \frac{1}{W} \sum_{j=1}^{W} \sum_{v \in \mathcal{V}} \pi(v \mid q, c_{i, <j}) \log \pi(v \mid q, c_{i,<j})
\end{equation}

This entropy term is then scaled and clipped before integration with the advantage function:
\begin{equation}
A_i^{\text{new}} = A_i + \min \left( \lambda E_i, \frac{|A_i|}{\alpha} \right),
\end{equation}
where $\lambda > 0$ is a scaling factor for entropy weighting, and $\alpha > 1$ governs the clipping threshold. The clipping via \(\frac{|A_i|}{\alpha}\) prevents the entropy adjustment from inverting the sign of the original advantage $A_i$, thereby maintaining the optimization direction while fostering diverse CoT explorations—particularly when $A_i$ is small~\cite{shao2024deepseekmath,schulman2015high,cheng2025reasoning}. This mechanism strikes a balance between directed learning and exploratory reasoning, enhancing the model's robustness in function calling tasks.

%% file: sections/experiment.tex
\section{Experiments}

In this section, we first introduce a simple RL data filtering on public datasets. Then, we elaborate on experimental settings and main results. Ablation study and a deep analysis are presented at the end.  

\subsection{Data Preparation}

\begin{figure}[t]
    \centering
    \includegraphics[width=1.0\linewidth]{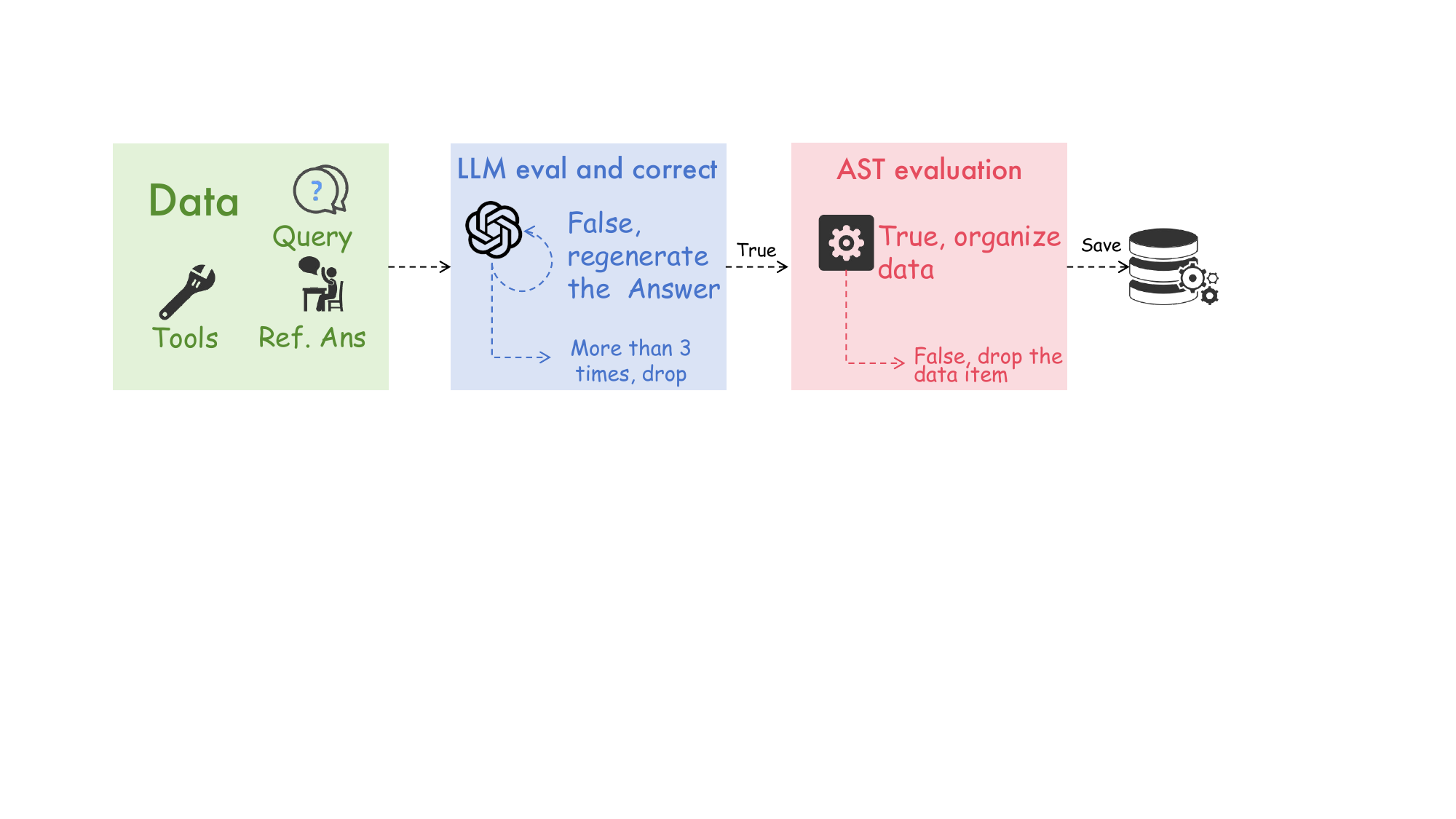}
    \caption{Implementation of the data cleaning pipeline for reinforcement learning in function calling. We begin with LLM-based evaluation and correction, followed by Abstract Syntax Tree (AST) evaluation. Data is retained only after passing all stages or discarded after three regeneration attempts.}
    \vspace{-4mm} 
    \label{fig:data}
\end{figure}

Our data preparation pipeline is engineered to refine high-quality samples for reinforcement learning in function calling tasks. For each initial input tuple $(q, T, g)$, $q$ denotes the user query, $T$ represents a set of available tools, and $g$ is the reference answer. The data undergoes a rigorous two-stage evaluation process, as depicted in Figure~\ref{fig:data}.
The pipeline consists of the following evaluations:

\begin{enumerate}
    \item \textbf{LLM Evaluation and Correction:} An initial assessment is conducted using GPT4-o~\cite{hurst2024gpt} to evaluate the query-tool-answer tuple. If discrepancies or errors are detected, the LLM regenerates the answer to rectify them. To maintain data quality, a strict dropout mechanism is enforced: samples are discarded if regeneration exceeds three attempts. This stage yields a binary outcome, denoted as $\operatorname{Eval}_{\mathrm{LLM}}(q, T, g) \in \{\mathrm{True, False}\}$.
    \item \textbf{Abstract Syntax Tree (AST) Evaluation:} Following a successful LLM evaluation, an AST-based assessment is performed on the reference answer. A sample is discarded if: (i) the query cannot be addressed by any tool in $T$, but $g$ parses as a valid tool call via AST; or (ii) the query requires a tool call, but $g$ is free-form text that fails AST parsing. This stage also produces a binary result, denoted as $\operatorname{Eval}_{\mathrm{AST}}(q, T, g) \in \{\mathrm{True, False}\}$. The details of the AST are shown in Appendix~\ref{AST}.
\end{enumerate}

Only samples that pass both evaluations (i.e., $\operatorname{Eval}_{\mathrm{AST}}(q, T, g)\land \operatorname{Eval}_{\mathrm{LLM}}(q, T, g)$) are retained in the database. For our experiments, we refine the xlam-function-calling-60k dataset~\cite{zhang2024xlam} and Open-Agentic-tool-use~\cite{Open-Agentic-tool-use} dataset. 

\begin{table*}[h]
\small
\renewcommand{\arraystretch}{1.0}
\begin{tabular}{@{}lccccccccccc@{}}
\toprule
&& \multicolumn{5}{c}{\textbf{Multi-Turn}}& \multicolumn{3}{c}{\textbf{Single-Turn}}\\
\cmidrule(lr){3-7}\cmidrule(lr){8-10}
\textbf{Models}&\begin{tabular}[c]{@{}c@{}}\textit{Parameter}\\ \textit{Counts}\end{tabular}
&\multicolumn{1}{c}{\textit{Overall}} &
  \multicolumn{1}{c}{\textit{Base}} &
  \multicolumn{1}{c}{\begin{tabular}[c]{@{}c@{}}\textit{Miss}\\ \textit{Func}\end{tabular}} &
  \multicolumn{1}{c}{\begin{tabular}[c]{@{}c@{}}\textit{Miss}\\ \textit{Param}\end{tabular}} &
  \multicolumn{1}{c}{\begin{tabular}[c]{@{}c@{}}\textit{Long}\\ \textit{Context}\end{tabular}}&
  \multicolumn{1}{c}{\textit{Overall}} &
  \multicolumn{1}{c}{\textit{Non-Live}} &
  \multicolumn{1}{c}{\textit{Live}} \\ 
\midrule
\textbf{GPT-5-2025-08-07}&/&28.50	&33.50	&29.50	&23.00	&28.00	&65.59&72.92	&58.25\\
\textbf{GPT-4o-2024-11-20}&/&42.50	&55.50	&34.50	&29.00	&\underline{51.00}&77.21	&83.88	&70.54\\
\textbf{Gemini-2.5-Pro}&/&25.00	&25.50	&26.00	&24.50	&24.00&74.50	&85.04	&63.95\\
\textbf{o3-2025-04-16}&/&38.38&44.00	&40.50	&31.50	&37.50&53.01	&39.98	&66.03	\\
\textbf{Amazon-Nova-Pro-v1:0}&/&34.75	&42.50	&24.50	&27.50	&44.50	&81.78    &85.25	&78.31\\
\textbf{Grok-4-0709}&/&36.12	&44.00	&31.00	&26.00	&43.50	&79.80&85.21	&74.39\\
\midrule
\textbf{Moonshotai-Kimi-K2-Inst}&1000B&41.25	&51.00	&\underline{43.00}	&31.00	&40.00	&80.80&84.02	&77.57\\
\textbf{DeepSeek-R1-0528}&671B &\underline{44.50}	&54.50	&41.00	&36.50	&46.00&78.22&75.73	&\underline{80.90}\\	
\textbf{Qwen3-235B-A22B-Inst-2507}&245B&39.62	&53.50	&34.50	&27.50	&43.00	&\underline{83.37}&\textbf{90.12}	&76.61	\\
\textbf{Llama-4-Maverick}&400B&17.88	&23.50	&18.00	&14.00	&16.00 &80.90 &88.15	&73.65\\
\textbf{Qwen3-32B}&32B&47.50	&53.00	&50.50	&\underline{40.00}	&46.50&84.21	&87.96	&80.46\\
\textbf{ToolACE-2-8B}&8B&37.00	&47.00	&31.00	&28.00	&42.00&82.54	&87.87	&77.20\\
\textbf{BitAgent-8B}&8B&37.75	&46.50	&37.50	&24.00	&43.00&81.71	&87.33	&76.09\\
\textbf{watt-tool-8B}&8B&37.88	&45.50	&39.00	&24.00	&43.00&81.71	&87.54	&75.87\\
\textbf{ToolACE-MT} &8B& 40.25& \underline{57.50} & 31.50 &34.00 & 38.00 &78.23& 84.94& 71.52  \\
\textbf{Qwen3-4B-Inst-2507} &4B &15.75	&19.00	&15.50	&12.50	&16.00 &78.19	&86.35	&70.02  \\
\midrule
\textbf{EGPO-4B}&4B&\textbf{56.25} &\textbf{63.50}& \textbf{54.50}& \textbf{48.00} &\textbf{59.00}&\textbf{84.92}&\underline{88.50}&\textbf{81.34}\\
\bottomrule
\end{tabular}
\caption{\label{tab:bfcl-overall} Performance on BFCL (last updated August 26, 2025), with all metrics calculated using the official script. The best result within each category is highlighted in \textbf{bold}. The second best results are \underline{underlined}.} 
\end{table*}

\begin{table}[h]
  \centering
    \begin{tabular}{lc}
        \toprule
        \textbf{Processing Stage} & \textbf{Samples} \\
        \midrule
        xLAM+Open-Agentic & 90,000 \\
        After LLM Evaluation & 89,641 \\
        After AST Evaluation & 88,759 \\
        \bottomrule
    \end{tabular}
\caption{Data statistics in two-stage data preparation.}
\vspace{-4mm} 
\label{tab:traindata}
\end{table}
\subsection{Experiment Setting}

\noindent\textbf{Training Data.} We conduct experiments by using xlam-function-calling-60k~\cite{zhang2024xlam} and Open-Agentic-tool-use~\cite{Open-Agentic-tool-use} datasets, the former provides 60,000 single-turn samples, and the latter provides 30,000 multi-turn samples. The first stage of data filtering is employed with GPT-4o~\cite{hurst2024gpt}, and the second stage is to conduct an objective evaluation based on AST. Table~\ref{tab:traindata} summarizes the data statistics after each processing stage.

\noindent\textbf{Models.} To evaluate EGPO's generalizability, we experiment with diverse base models: Llama-3.2-3B-Instruct~\cite{grattafiori2024llama} from Llama family and Qwen3-4B-Instruct-2507 from Qwen family. In experiments, we compare against strong open-sourced specialized tool-calling models like ToolACE-2-8B~\cite{liu2024toolace}, BitAgent-8B, ToolACE-MT~\cite{zeng2025toolace}, and watt-tool-8B. We also include general-purposed models like GPT-4o~\cite{hurst2024gpt}, Gemini-2.5-Pro~\cite{comanici2025gemini}, and Grok-4-0709 as strong baselines.

\noindent\textbf{Benchmarks.} We evaluate both Single-Turn and Multi-Turn tool calling performance on the Berkeley Function Calling Leaderboard (BFCLv3)~\cite{patil2024gorilla}. Our evaluation is reported in terms of accuracy (\%), which is calculated using the official script. To further evaluate our model, we also report the results on more function call specialized benchmarks, ACEBench~\cite{chen2025acebench} and APIBank~\cite{li2023api}.

\noindent\textbf{Implementation Details.} We train models for 5 epochs with a learning rate of $1 \times 10^{-6}$ and inference temperature of 0.7 by using the Verl framework~\cite{sheng2025hybridflow}. Training is performed on 32 H200 GPUs with a batch size of 1,024, 8 rollouts and reserving 5\% of data for validation. We set the KL coefficient to 0.001 and the maximum response length to 8192 tokens. The hyperparameters $\lambda$ and $\alpha$ are set to 0.4 and 2.

\subsection{Experimental Results}

\noindent\textbf{Results on BFCL.} We present the detailed experimental results in Table~\ref{tab:bfcl-overall}. In the critical Multi-Turn evaluation, which requires complex state tracking and error recovery, EGPO-4B achieves the best performance. Our model, with only 4 billion parameters, attains an Overall score of 56.25\%, substantially outperforming all other evaluated models, including major proprietary models like GPT-4o (42.50\%) and Grok-4 (36.12\%). When compared to similarly sized open-source models (e.g., ToolACE-MT~\cite{zeng2025toolace} with 8B parameters, scoring 40.25\%), EGPO-4B demonstrates a remarkable 15.0 point advantage, validating the efficacy of the EGPO approach. Its superior performance is consistent across challenging sub-metrics. 

Notably, it leads in the two difficult error recovery scenarios: 1) Miss Func (missing certain required functions, 54.50\%). This demonstrates our model's ability to identify when no function is available to satisfy the user's request and to request additional tools. 2) Miss Param (missing certain required parameters, 48.00\%). This shows that the model can detect missing key information in a user's request and request clarification rather than making unfounded assumptions, which is crucial for real-world robustness. Furthermore, EGPO also achieves the highest score in the Long Context metric (59.00\%), indicating that EGPO effectively improves the model’s ability to maintain coherence and leverage historical context in extended conversations. For the Base test, EGPO-4B also secures the top result at 63.50\%, confirming a strong foundational capability. 

In the Single-Turn evaluation, which measures immediate response accuracy and precision, EGPO-4B maintains its competitive standing. The model secures the best Overall Single-Turn accuracy among all evaluated models, achieving 84.92\%. This performance surpasses near large open-sourced competitor, Qwen3-235B (83.37\%), and far exceeds the strong baselines like GPT-4o (77.21\%). This result highlights that the EGPO approach enhances Multi-Turn function call skills without sacrificing the core Single-Turn precision.

These results demonstrate that the EGPO-4B’s parameter efficiency among LLMs. By achieving the highest Overall score in both the complex Multi-Turn evaluation (56.25\%) and the challenging Single-Turn evaluation (84.92\%), our method successfully bridges the performance gap between small, resource-efficient models and large-scale proprietary or publicly available models, even those orders of magnitude larger in parameter count.

\begin{figure*}[!t]
    \centering
    \begin{subfigure}[b]{0.49\textwidth}
        \centering
        \includegraphics[width=\textwidth]{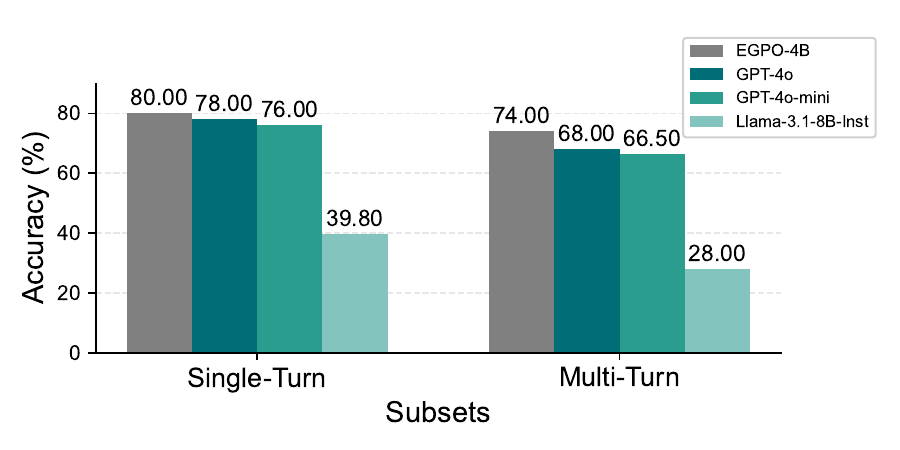}
        \caption{Model performance on ACEBench.}
        \label{fig:left}
    \end{subfigure}
    \hfill
    \begin{subfigure}[b]{0.49\textwidth}
        \centering
        \includegraphics[width=\textwidth]{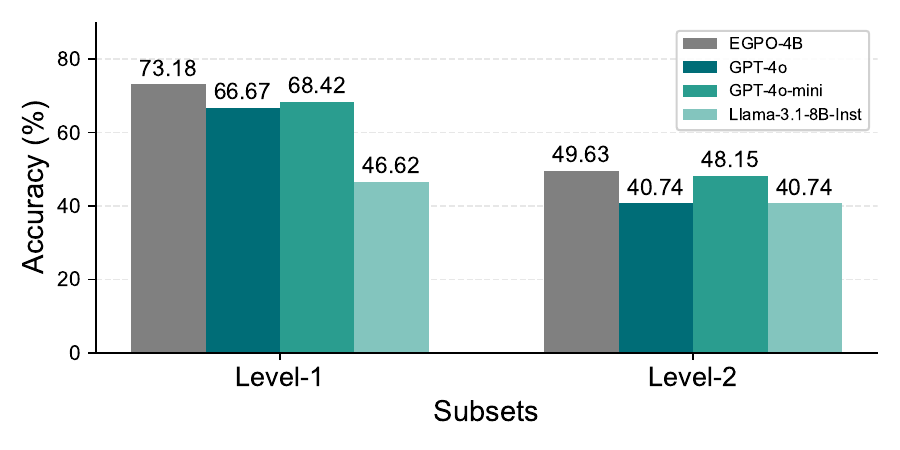}
        \caption{Model performance on APIBank.}
        \label{fig:right}
    \end{subfigure}
    \caption{Performance on ACEBench and  APIBank with all metrics calculated using the official scripts.}
    \label{tab:model_performance}
\end{figure*}

\noindent\textbf{Results on ACEBench and APIBank.} Figure~\ref{fig:left} illustrates the models' performance on the ACEBench Benchmark, including Single-Turn and Multi-Turn subsets. In the Single-Turn tasks, the EGPO-4B model achieves the highest accuracy at 80.00\%, demonstrating a slight edge over other models, which are closely clustered. GPT-4o scores 78.00\% and GPT-4o-mini scores 76.00\%. The superiority of EGPO-4B is more pronounced in the Multi-Turn subset, which requires sophisticated conversational coherence. EGPO-4B leads with an accuracy of 74.00\%, creating a larger performance gap over GPT-4o (68.00\%) and GPT-4o-mini (66.50\%). Figure~\ref{fig:right} presents the results on the APIBank Benchmark, which evaluates function-calling capabilities and is segmented into Level-1 (Call) and Level-2 (Retrieve+Call). In the Level-1 tasks, EGPO again exhibits the top performance with 73.18\% accuracy. GPT-4o and GPT-4o-mini are competitive, achieving 66.67\% and 68.42\%, respectively. Llama-3.1-8B-Instruct is noticeably lower at 46.62\%. In the more demanding Level-2 subset, which involves more complex function-calling scenarios, all models experience a decline in accuracy, but EGPO maintains its leading position with 49.63\%. Intriguingly, on this harder subset, GPT-4o and GPT-4o-mini score identically at 40.74\%. Our results show that models trained via the proposed EGPO method generalize well across different tool-calling scenarios, unlocking new chances to improve performance scaling in LLM tool-augmented learning.

\begin{figure*}[!t]
    \centering
    \begin{subfigure}[b]{0.244\textwidth}
        \centering
        \includegraphics[width=\textwidth]{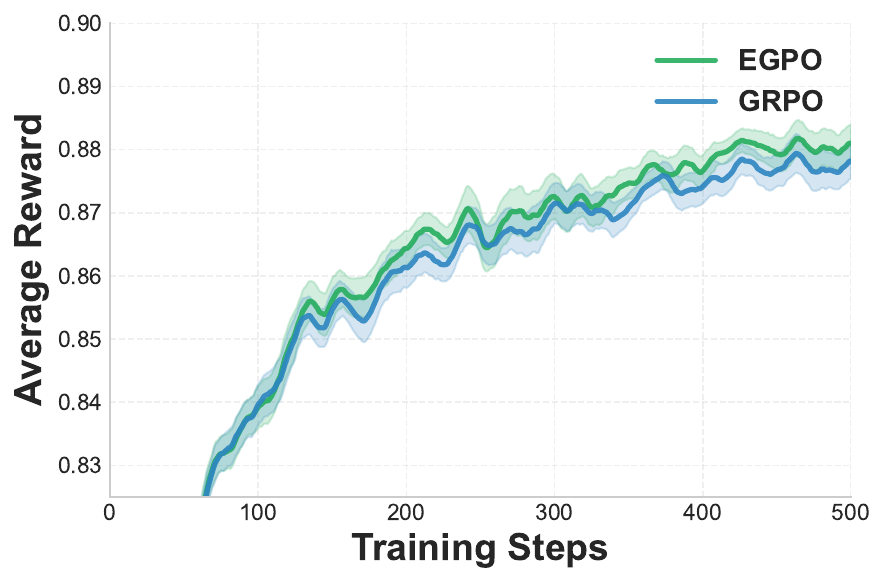}
        \caption{Average Reward.}
        \label{fig:a}
    \end{subfigure}
    \hfill
    \begin{subfigure}[b]{0.244\textwidth}
        \centering
        \includegraphics[width=\textwidth]{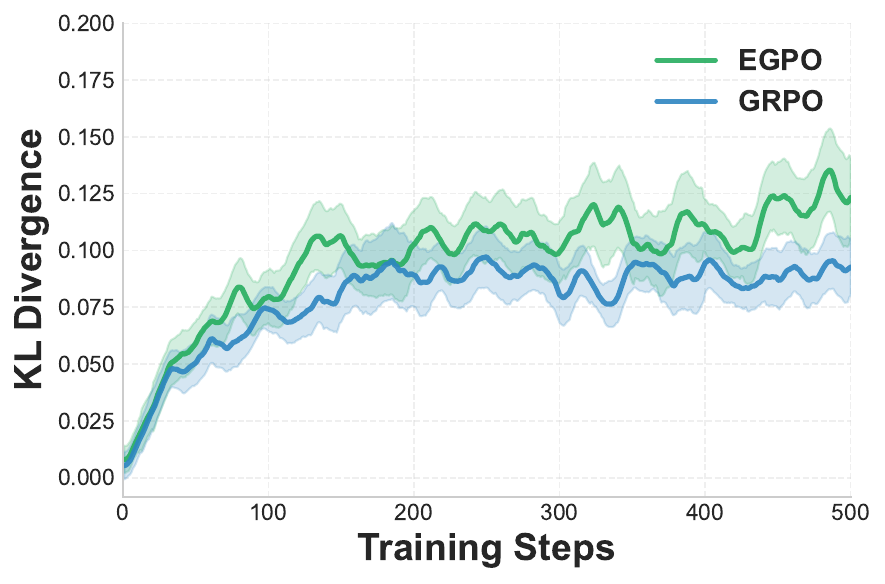}
        \caption{KL Divergence.}
        \label{fig:b}
    \end{subfigure}
    \hfill
    \begin{subfigure}[b]{0.244\textwidth}
    \centering
    \includegraphics[width=\textwidth]{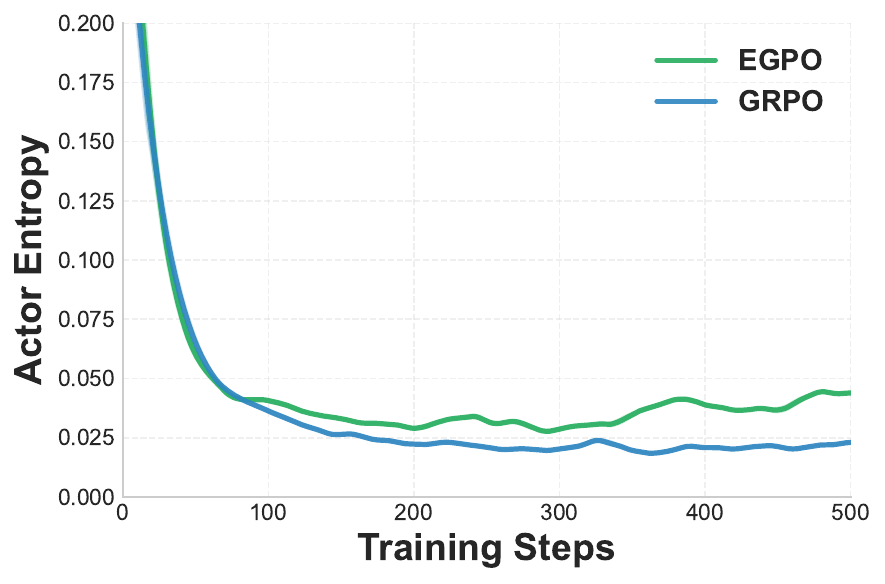}
    \caption{Actor Entropy.}
    \label{fig:c}
\end{subfigure}
\hfill
\begin{subfigure}[b]{0.244\textwidth}
    \centering
    \includegraphics[width=\textwidth]{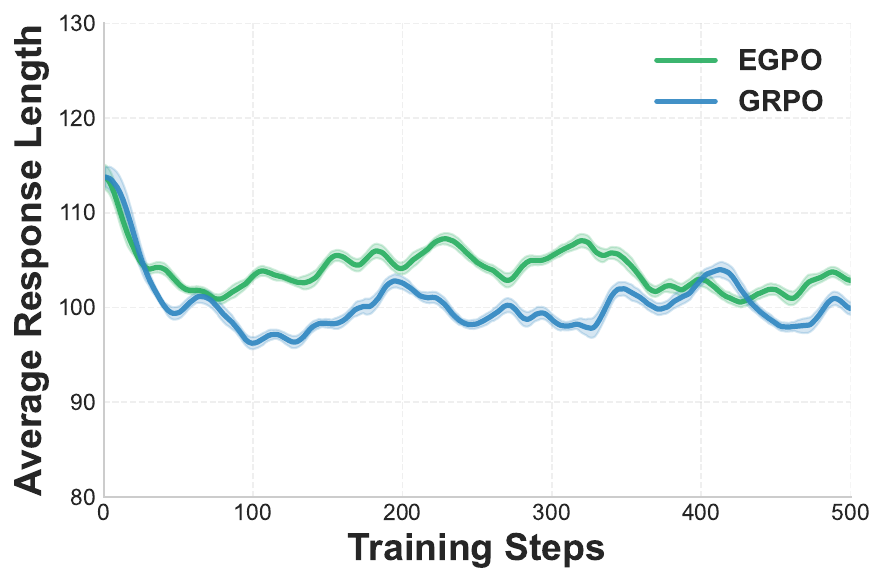}
    \caption{Average Response Length.}
    \label{fig:d}
\end{subfigure}
    \caption{Visualization of the learning curves for EGPO and GRPO during training. We report the Average Reward, KL Divergence, Actor Entropy and Average Response Length.}
    \label{fig:entropy}
\end{figure*}

\subsection{Ablation Study}
\begin{figure*}[!t]
    \centering
    \begin{subfigure}[b]{0.49\textwidth}
        \centering
        \includegraphics[width=\textwidth]{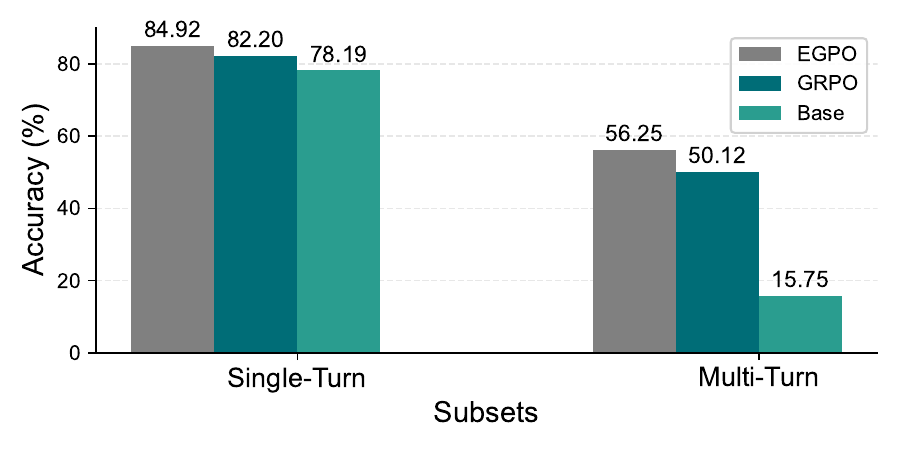}
        \caption{Performance of Qwen3-4B-Instruct-2507.}
        \label{fig:4b}
    \end{subfigure}
    \hfill
    \begin{subfigure}[b]{0.49\textwidth}
        \centering
        \includegraphics[width=\textwidth]{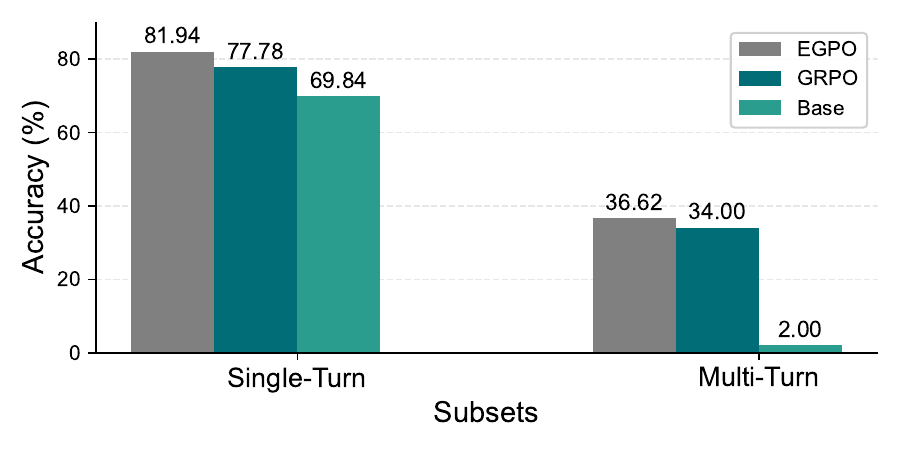}
        \caption{Performance of Llama3.2-3B-Instruct.}
        \label{fig:3b}
    \end{subfigure}
    \caption{Performance of two series models trained by EGPO and GRPO on BFCL. }
    \label{tab:3B and 4B}
\end{figure*}

In the ablation study, we adopt consistent configurations across all experiments to ensure fair comparisons and select the GRPO method as the baseline to evaluate the performance of each module. 

\noindent\textbf{Comparison of EGPO and GRPO.} In Figure~\ref{tab:3B and 4B},  we present a comparative analysis by using two different models. After training models by using the EGPO and GRPO with the same Single-Criteria Reward, evaluated on the Single-Turn and Multi-Turn subsets of BFCL. Across both the Qwen3-4B model and Llama3.2-3B model, the EGPO approach consistently yields the highest accuracy in all four sub-categories, demonstrating its effectiveness and generalizability. Crucially, the advantage of EGPO is most pronounced in the difficult Multi-Turn tasks. For the Qwen model, EGPO scores 56.25\% compared to the Base model's mere 15.75\%, and for the Llama model, EGPO achieves 36.62\% compared to the Base model's collapse at just 2.00\%, underscoring that entropy from the CoT provides a dramatic and vital boost in handling complex, sequential Multi-Turn conversations.

\noindent\textbf{Training Dynamics.} We present the visualized Qwen3-4B learning curves in Figure~\ref{fig:entropy}, including Average Reward,  KL Divergence, Actor Entropy, and Average Response Length.  During the training steps, the reward curve in Figure~\ref{fig:a} of EGPO is consistently above that of GRPO, which demonstrates the training stability and excellent performance of EGPO. During the middle and later stages of training in Figure~\ref{fig:b}, the KL divergence of EGPO is significantly higher than that of GRPO, deviating further from the base model, implying that introducing the entropy of the CoT in the advantage computation allows the model to better explore a thinking pattern suitable for function calling scenarios. From the perspective of the Actor Entropy in Figure~\ref{fig:c}, unlike GRPO, where the entropy consistently decreases and then plateaus. EGPO’s curve first declines and then rises, according to~\cite{yu2025dapo}, the entropy of the actor model with such a trend change is associated with the model’s exploration capability, maintaining a trend of slow entropy increase is conducive to the model generating diverse answers and achieving improved performance. Finally, from the perspective of response length in Figure~\ref{fig:d}, the responses generated by EGPO are longer than those by GRPO. This additional token budget enables the model to produce responses with better quality.

\noindent\textbf{Study of Reward Patterns.} As shown in Table~\ref{tab:rewardcompare}, we also use two series of models for reward pattern exploration and find that reward patterns with Reasoning Format often fail to achieve optimal results. Here, we set up an additional reward of 0.5 if the model outputs the correct Reasoning Format (i.e., enclosing the CoT within <think> and </think> tags), but in our experiments, this reward pattern does not yield the best results. This suggests that for the models with instruction-following capability, format reward setting will cause reward redundancy and fail to achieve optimal results.

\begin{table}[htbp]
\centering
\small
\begin{tabular}{@{}lcccc@{}}
\toprule
\multirow{2}{*}{\textbf{Subsets}} & \multicolumn{2}{c}{\textbf{Qwen3-4B}} & \multicolumn{2}{c}{\textbf{Llama-3-3B}}\\ 
\cmidrule(lr){2-3}\cmidrule(lr){4-5}
& \textit{w/ RF} & \textit{w/o RF} & \textit{w/ RF} &\textit{w/o RF} \\
\midrule
Single-Turn & 84.36 & \textbf{84.92} &  77.69& \textbf{81.94} \\
Multi-Turn & 53.25 & \textbf{56.25} & 32.50 & \textbf{36.62} \\
\bottomrule
\end{tabular}
\caption{Performance comparison of different models with or without Reason Format reward.}
\label{tab:rewardcompare}
\end{table}

\subsection{Further Analysis}

\begin{figure}[tbp]
  \centering
  \includegraphics[width=\columnwidth]{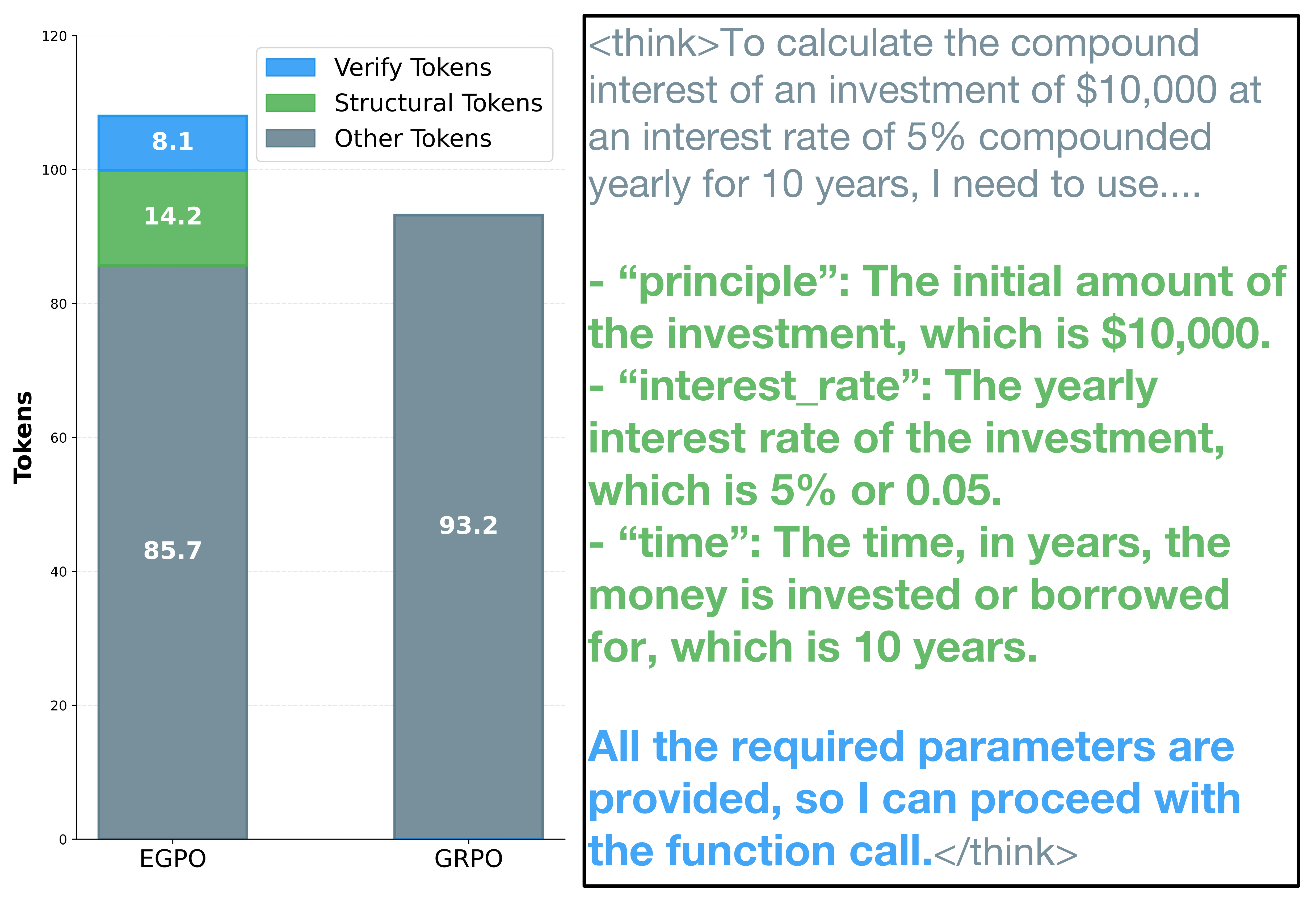}
  \caption{Statistical results of the special action tokens generated by EGPO.}
  \label{tab:tokenststis} 
\end{figure}

In Figure~\ref{fig:d}, we observe that EGPO-trained models exhibit longer CoT compared to GRPO. To further investigate this, we randomly sampled 100 pieces from the BFCL simple Python subset and compared EGPO with GRPO. Unlike GRPO, the EGPO-trained model self-generates two types of special actions: structural parameter checks, used to verify parameter correctness, and verification, used for validation of the final answer. As shown in Figure~\ref{tab:tokenststis}, these action tokens account for over 20\% of the total tokens in the CoT, suggesting that EGPO induces the model to adopt a distinct thinking pattern. This behavior is likely driven by the entropy reward in EGPO, which encourages more exploration, resulting in structural and verification actions that enhance overall model performance.


%% file: sections/appendix.tex
\section{Q \& A Template}

The following Q \& A Template in Figure~\ref{QATemplate}, with its predefined set of instructions, guidelines, and constraints, plays a pivotal role in standardizing the behavior, output format of the model being trained and evaluated. It ensures that during the training phase, the model learns in a consistent manner aligned with the desired objectives, and during the evaluation, the performance is generated under a unified set of rules.

\begin{figure*}
\begin{center}
\begin{tcolorbox}[title={Q \& A Template},colframe=blue!70]
\textcolor{red}{System Prompt}\\
You are an expert in composing functions. You are given a question and a set of possible functions. Based on the question, you will need to make one or more function/tool calls to achieve the purpose. If none of the function can be used, point it out. If the given question lacks the parameters required by the function, also point it out. You should only return the function call in tools call sections.\\
\# Tool\\
Here is a list of functions in JSON format that you can invoke:\\
\{\texttt{functions}\}\\
In each action step, you MUST:\\
1. Think about the reasoning process in the mind and enclosed your reasoning within \textless{}think\textgreater{}\textless{}/think\textgreater{} XML tags.\\
2. Then, provide a function call with function names and arguments. i.e.,\begin{sloppy}
[func\_name1(params\_name1=params\_value1, params\_name2=params\_value2...), func\_name2(params)]
\end{sloppy}\\
3. Make sure both the reasoning and the tool call steps are included together in one single reply.\\
A complete reply example is: \\
\textless{}think\textgreater{}reasoning process here\textless{}/think\textgreater{}\\
answer here\\
Please make sure the type of the arguments is correct and reject to answer if the function can not solve the user's question.\\

\textcolor{orange}{User Query}\\
Please ...\\

\textcolor{blue}{Answer}\\
\textless{}think\textgreater{}reasoning process here\textless{}/think\textgreater{}\\
answer here\\

\end{tcolorbox}
\end{center}
\caption{Q \& A Template used in the experiments.}
\label{QATemplate}
\end{figure*}
\section{Dataset}
\label{Dataset}

\subsection{Data Source}

The single-turn part is from the xlam-function-calling-60k~\cite{zhang2024xlam} and the multi-turn part comes from the Open-Agentic-tool-use~\cite{Open-Agentic-tool-use}. The Open-Agentic-tool-use dataset is an open-source project focusing on multi-turn tool-use learning and serving as a reproduction of Kimi K2’s Large-Scale Agentic Data Synthesis for Tool Use Learning~\cite{team2025kimik2}. Built based on 8 environments, it currently has over 30000 open-sourced data and is suitable for both SFT and RL training. Its data generation process involves constructing API relationship graphs for environments, sampling APIs with random parameter construction, verifying API executability in the environment, generating user questions via a user agent based on sampled APIs and environment feedback. The xlam-function-calling-60k is a high-quality, verifiable synthetic text dataset designed for training and evaluating LLM function-calling capabilities. Comprising 60,000 entries, it includes 33,659 entries generated by DeepSeek-V2-Chat and the rest by Mixtral-8x22B-Inst, leveraging 3,673 executable APIs across 21 categories; each entry undergoes three-stage verification (format checking, function execution, semantic validation), resulting in over 95\% accuracy in human evaluation (with minor issues like inaccurate arguments in the remaining 5\%). Structurally, each JSON entry links a plain-language query (user request), an array of tools (available APIs with names, descriptions, and parameter details like type/required status), and an answers array (correct API calls + arguments). It supports tasks like question answering, text generation, and reinforcement learning for LLM Agents.


\section{Full Results on Leader Boards}

\subsection{Single-Turn Results on BFCL}

Here, we show all models' full Single-Turn results in Table~\ref{tab2}, where EGPO-4B attains superior performance with scores of 88.50\% and 81.34\% on Non-Live and Live sections. The results surpass all similar scale models and achieve great enhancement on Parallel and Parallel Multiple subsections, which indicates that our model has achieved excellent tool-calling capabilities under complex single-turn queries.

\begin{table*}[thb]
\scriptsize
\centering
\begin{tabular}{@{}lcccccccccc@{}}
\toprule

& \multicolumn{5}{c}{\textbf{Non-Live}} & \multicolumn{5}{c}{\textbf{Live}}  \\
\cmidrule(lr){2-6} \cmidrule(lr){7-11}
\textbf{Models} &Overall &Simple & Multiple & Parallel & \begin{tabular}[c]{@{}c@{}}Parallel\\ Multiple\end{tabular} &Overall& Simple & Multiple & Parallel & \begin{tabular}[c]{@{}c@{}}Parallel\\ Multiple\end{tabular} \\
\midrule
\multicolumn{11}{c}{Close-Sourced Models} \\
\midrule
\textbf{GPT-5-2025-08-07}&72.92	&58.67	&76.00	&84.00	&73.00	&58.25	&61.63	&57.45	&50.00	&62.5\\
\textbf{GPT-4o-2024-11-20}&83.88	&76.5	&91.00	&90.00	&78.00	&70.54	&70.54	&70.75	&62.50	&66.67\\
\textbf{Gemini-2.5-Pro} &85.04	&68.67	&91.00	&91.50	&89.00	&63.95	&75.58	&60.40	&81.25	&83.33\\
\textbf{o3-2025-04-16}&39.98	&71.42	&88.50	&0.00	&0.00	&66.03	&69.77	&67.62	&0.00	&0.00	\\
\textbf{Amazon-Nova-Pro-v1:0}&85.25	&71.50	&93.50	&92.00	&84.00	&78.31	&80.62	&77.87	&87.50	&66.67\\
\textbf{Grok-4-0709}&85.21	&73.33	&92.50	&89.50	&85.50	&74.39	&77.91	&73.69	&68.75	&70.83\\
  \midrule
\multicolumn{11}{c}{Open-Sourced Models} \\
\midrule
\textbf{Moonshotai-Kimi-K2-Inst}&84.02	&79.08	&93.00	&76.00	&88.00	&77.57	&86.05	&75.78	&81.25	&62.50\\
\textbf{DeepSeek-R1-0528}&86.52	&71.58	&96.00	&91.00	&87.50	&77.65	&83.72	&76.16	&87.50	&70.83\\	
\textbf{Qwen3-235B-A22B-Inst-2507}&90.12	&80.50	&95.00	&95.50	&89.50	&76.61	&83.72	&75.02	&75.00	&70.83	\\
\textbf{Llama-4-Maverick}&88.15	&76.08	&95.00	&92.50	&89.00	&73.65	&84.50	&71.04	&75.00	&70.83\\
\textbf{Qwen3-32B}&87.96	&76.83	&94.50	&90.00	&90.50	&80.46	&84.88	&79.87	&81.25	&58.33\\	
\textbf{ToolACE-2-8B}&87.87	&75.00	&92.00	&92.50	&92.00	&77.20	&70.16	&79.30	&75.00	&62.5\\
\textbf{BitAgent-8B}&87.33	&76.33	&95.00	&93.00	&85.00	&76.09	&78.29	&75.59	&87.50	&66.67\\
\textbf{watt-tool-8B}&87.54	&76.17	&94.50	&94.00	&85.50	&75.87	&77.52	&75.59	&87.50	&62.50\\
\midrule
\multicolumn{11}{c}{Base Models} \\
\midrule
\textbf{Qwen3-4B-Instruct-2507}&86.35	&76.42	&91.50	&88.00	&89.50	&70.02	&73.64	&69.61	&81.25	&41.67\\
\textbf{Llama-3.2-3B-Inst}&81.94	&71.25	&92.00	&87.00	&77.50	&57.74	&63.57	&57.26	&25.00	&37.50\\
\midrule
\multicolumn{11}{c}{Ours} \\
\midrule
\textbf{EGPO-4B} &88.50&76.50&95.00&93.50&89.00&81.34&85.66&79.49&87.50&75.00 \\
\textbf{EGPO-3B} &87.38&72.00&96.00&89.00&92.50&76.50&74.42&75.12&81.25&70.83 \\
\bottomrule
\end{tabular}%

\caption{The evaluation details on the BFCL benchmark (last updated August 26, 2025), with all metrics calculated using the official script. }
\label{tab2}
\end{table*}

\subsection{ACEBench}

\begin{table*}[h]
\centering
\begin{tabular}{lccccc}
\toprule
\textbf{Models} & \textbf{Atom} & \textbf{Single-Turn} & \textbf{Multi-Turn} & \textbf{Similar API} & \textbf{Preference}  \\
\midrule
\textbf{GPT-4o-2024-11-20} & \textbf{90.0} & 78.0 & 68.0 & \textbf{80.0} & \textbf{78.0}  \\
\textbf{Llama3.1-70B-Inst} & 83.7 & 71.5 & 61.0 & 74.0 & 66.0  \\
\textbf{ToolACE-MT-8B} & 83.0 & 64.0 & 51.0 & 68.0 & 68.0 \\
\midrule
\multicolumn{6}{c}{Base Models} \\
\midrule
\textbf{Qwen3-4B-Instruct-2507}&67.7 &60.0 &52.0 &68.0&64.0\\
\textbf{Llama3.2-3B-Inst} & 27.0 & 19.0 & 7.0 & 38.0 & 30.0 \\
\midrule
\multicolumn{6}{c}{Ours} \\
\midrule
\textbf{EGPO-4B} & 83.0 & \textbf{80.0} & \textbf{74.0} & \textbf{80.0} & 50.0  \\
\textbf{EGPO-3B} & 66.3 & 40.0 & 42.0 & 58.0 & 30.0 \\
\bottomrule
\end{tabular}
\caption{Full result of the accuracy of ACEBench normal examples. The best result within each category is highlighted in \textbf{bold}.}
\label{ACEfull}
\end{table*}

Table~\ref{ACEfull} presents the comprehensive results of various models on the ACEBench normal metric, encompassing multiple evaluation dimensions. For the Single-Turn task, EGPO-4B stands out with the highest accuracy of 80.0\%. In the Multi-Turn task, EGPO-4B again performs best, obtaining an accuracy of 74.0\%. 

\subsection{Results on BFCLv4}

BFCLv4 evaluates the model's agentic capability with two sections, Web Search and Memory. We test our model, EGPO-4B, on the two corresponding test sets in Table~\ref{tab:bfclv4}. We do not use any datasets related to search or memory, but still improve the scores in both sections by 10.5\% and 5.16\%.

\begin{table*}[h]
\small
\centering
\begin{tabular}{@{}lccccccc@{}}
\toprule
&\multicolumn{3}{c}{\textbf{Web Search}}& \multicolumn{4}{c}{\textbf{Memory}}\\
\cmidrule(lr){2-4}\cmidrule(lr){5-8}
\textbf{Models}
&\multicolumn{1}{c}{\textit{Overall}} &
  \multicolumn{1}{c}{\textit{Base}} &
  \multicolumn{1}{c}{ \textit{No Snippet}} &
  \multicolumn{1}{c}{\textit{Overall}} &
  \multicolumn{1}{c}{\textit{KV}} &
  \multicolumn{1}{c}{\textit{Vector}}&
  \multicolumn{1}{c}{\textit{Recursive Sum}} \\ 
  \midrule
\textbf{Qwen3-4B-Instruct-2507} &5.00	&4.00	&6.00	&12.69	&11.61	&11.61	&14.84\\
\textbf{EGPO-4B} &15.50&15.00&16.00&17.85&8.39&0.65&44.52\\
\bottomrule
\end{tabular}
\caption{\label{tab:bfclv4} Performance of EGPO-4B on BFCLv4.}
\end{table*}

\section{Abstract Syntax Tree Parser}
\label{AST}

The Abstract Syntax Tree (AST) parser is a foundational component of a compiler or interpreter, responsible for transforming the linear stream of tokens from a lexer into a tree-like, hierarchical representation of the program's structure. This AST is crucial because it abstracts away details like punctuation and whitespace, providing a clean, logical map of code elements that is far easier for later stages—such as optimization and code generation—to process than raw text. The function call evaluation process uses this AST in two phases: first, Function Matching strictly verifies the function name and parameter signature, enforcing specific rules for data types (e.g., direct boolean matching, order-dependent list matching), with multi-function checks using an all-or-nothing principle; second, Executable Evaluation verifies function output, checking non-REST tests for completeness or structural criteria, and REST tests for API validity and response structure, also strictly adhering to the all-or-nothing rule for multi-function execution. A simple example of Abstract Syntax Tree Parser is shown in Figure~\ref{ASTParserExample}.

\begin{figure*}
\begin{center}
\begin{tcolorbox}[
    title={AST Parser Example},
    colframe=blue!70,
    boxsep=5pt, 
    arc=4pt, 
    auto outer arc,
    coltitle=white, 
    colbacktitle=blue!70
]

\textbf{Example:}
\begin{verbatim}
[calculate_triangle_area(base=10, height=5)]
\end{verbatim}
\textbf{Parse:}
\begin{lstlisting}
Module(body=[
    Expr(value=List(elts=[
        Call(
            func=Name(id='calculate_triangle_area', ctx=Load()), 
            args=[], 
            keywords=[
                keyword(arg='base', value=Constant(value=10)), 
                keyword(arg='height', value=Constant(value=5))
                ]
            )
        ], ctx=Load()))
    ], type_ignores=[])
\end{lstlisting}

\end{tcolorbox}
\end{center}
\caption{AST Parser Example.}
\label{ASTParserExample}
\end{figure*}

\section{Case Study}

We further show some cases where the EGPO model uses additional special action tokens to assist in reasoning during the inference process in Figure~\ref{fig:case_study1} and Figure~\ref{fig:case_study2}. We refer to the tokens related to structural output as structural tokens and mark them in green, while those related to verification are called verify tokens and marked in blue. These special action tokens indicate that our model is executing tasks with new thinking patterns in tool-using scenarios; that is, EGPO helps the model explore better domain-specific thinking patterns.

\begin{figure*}[thb]
\centering
\begin{tcolorbox}[
  colback=gray!5!white,
  title={Case Study 1},
  boxrule=0.3mm,
  width=0.95\textwidth,
  arc=3mm,
  auto outer arc=true,
  fontupper=\small
]
\textcolor{red}{\textbf{System:}} You are an expert in composing functions. You are given a question and a set of possible functions. Based on the question, you will need to make one or more function/tool calls to achieve the purpose. If none of the function can be used, point it out. If the given question lacks the parameters required by the function, also point it out. You should only return the function call in tools call sections.\\
\# Tool\\
\medskip
\textbf{Here is a list of functions in JSON format that you can invoke:}
\begin{scriptsize}
\begin{verbatim}
[{
  "name": "calculate_compound_interest",
  "description": "Calculates the compound interest of an 
                  investment over a given time period.",
  "parameters": {
    "type": "dict",
    "properties": {
      "principle": {
        "type": "integer",
        "description": "The initial amount of the investment."
      },
      "interest_rate": {
        "type": "float",
        "description": "The yearly interest rate of the 
                        investment."
      },
      "time": {
        "type": "integer",
        "description": "The time, in years, the money is 
                        invested or borrowed for."
      },
      "compounds_per_year": {
        "type": "integer",
        "description": "The number of times the interest is 
                        compounded per year. Default is 1."
      }
    },
    "required": ["principle", "interest_rate", "time"]
  }
}]
\end{verbatim}
\end{scriptsize}
In each action step, you MUST:...\\
\medskip
\textcolor{orange}{\textbf{User:}} Calculate the compound interest of an investment of \$10,000 at an interest rate of 5\% compounded yearly for 10 years.

\medskip
\textcolor{blue}{\textbf{Assistant:}} \texttt{<think>} To calculate the compound interest of an investment of \$10,000 at an interest rate of 5\% compounded yearly for 10 years, I need to use the \texttt{calculate\_compound\_interest} function. The function requires the following parameters:
\begin{itemize}\color{OliveGreen}
\item \texttt{principle}: The initial amount of the investment, which is \$10,000.
\item \texttt{interest\_rate}: The yearly interest rate of the investment, which is 5\% or 0.05.
\item \texttt{time}: The time, in years, the money is invested or borrowed for, which is 10 years.
\item \texttt{compounds\_per\_year}: The number of times the interest is compounded per year, which is 1 since it is compounded yearly.
\end{itemize}
{\color{RoyalBlue}All the required parameters are provided, so I can proceed with the function call.} \texttt{</think>}
\medskip
\texttt{[calculate\_compound\_interest(principle=10000, interest\_rate=0.05, time=10, compounds\_per\_year=1)]}
\end{tcolorbox}
\caption{Example 1 of EGPO.}
\label{fig:case_study1}
\end{figure*}

\begin{figure*}[thb]
\centering
\begin{tcolorbox}[
  colback=gray!5!white,
  title={Case Study 2},
  boxrule=0.3mm,
  width=0.95\textwidth,
  arc=3mm,
  auto outer arc=true,
  fontupper=\small
]

\textcolor{red}{\textbf{System:}} You are an expert in composing functions. You are given a question and a set of possible functions. Based on the question, you will need to make one or more function/tool calls to achieve the purpose. If none of the functions can be used, point it out. If the given question lacks the parameters required by the function, also point it out. You should only return the function calls in your response. If you decide to invoke any of the function(s), you MUST put it in the format of \texttt{[func\_name1(params\_name1=params\_value1, params\_name2=params\_value2...), func\_name2(params)]}.

You SHOULD NOT include any other text in the response.

\medskip
\textbf{Here is a list of functions in JSON format that you can invoke:}

\begin{scriptsize}
\begin{verbatim}
[{
  "name": "convert_currency",
  "description": "Converts an amount from a particular currency 
                  to another currency.",
  "parameters": {
    "type": "dict",
    "properties": {
      "base_currency": {
        "type": "string",
        "description": "The base currency in which the original 
                        amount is present."
      },
      "target_currency": {
        "type": "string",
        "description": "The currency to which you want to 
                        convert."
      },
      "amount": {
        "type": "integer",
        "description": "The amount you want to convert."
      }
    },
    "required": ["base_currency", "target_currency", "amount"]
  }
}]
\end{verbatim}
\end{scriptsize}
In each action step, you MUST:...\\
\medskip
\textcolor{orange}{\textbf{User:}} How much will 20000 Japanese Yen be in United States Dollar?\\
\medskip
\textcolor{blue}{\textbf{Assistant:}} \texttt{<think>} To convert 20000 Japanese Yen to United States Dollar, I need to use the \texttt{convert\_currency} function. The function requires the base currency, target currency, and the amount to be converted. Here are the values I have:
\begin{itemize}\color{OliveGreen}
\item Base currency: Japanese Yen (JPY)
\item Target currency: United States Dollar (USD)
\item Amount: 20000
\end{itemize}
{\color{RoyalBlue}All the required parameters are provided, so I can proceed with the function call.} \texttt{</think>}
\medskip
\texttt{[convert\_currency(base\_currency='JPY', target\_currency='USD', amount=20000)]}
\end{tcolorbox}
\caption{Example 2 of EGPO.}
\label{fig:case_study2}
\end{figure*}